Original Article

# Weakly Supervised Teacher–Student Framework with Progressive Pseudo-mask Refinement for Gland Segmentation

## Running Title: Progressive Pseudo-Masks for Gland Segmentation

Hikmat Khan*, Wei Chen and Muhammad Khalid Khan Niazi

Department of Pathology, College of Medicine, The Ohio State University Wexner Medical Center, Columbus, OH, USA

***Correspondence to:** Hikmat Khan, Department of Pathology, College of Medicine, The Ohio State University Wexner Medical Center, Pelotonia Research Center, Columbus, OH 43210, USA. ORCID: 0009-0008-2550-1991 Tel: +1-856-405-XXXX, E-mail: Hikmat.khan@osumc.edu | Hikmat.khan179@gmail.com

# Abstract

**Background and objectives**: Colorectal cancer histopathological grading relies on the accurate segmentation of glandular structures. Current deep learning–based methods depend heavily on large-scale pixel-level annotations that are labor-intensive and not amenable to clinical practice. Weakly supervised semantic segmentation offers a promising alternative; yet existing class activation map–based weakly supervised semantic segmentation approaches often produce incomplete, low-quality pseudo-masks that overemphasize discriminative regions and fail to provide reliable supervision for unannotated glandular structures, limiting their suitability for dense histopathology segmentation under sparse supervision. We propose a novel weakly supervised teacher–student framework that leverages sparse pathologists' annotations and an Exponential Moving Average–stabilized teacher network to generate refined pseudo-masks.

**Methods:** Our framework integrates confidence-based filtering, adaptive fusion of teacher predictions with limited ground truth, and curriculum-guided refinement, enabling the student network to progressively delineate and accurately segment unannotated glandular regions. We validated our framework on an institutional colorectal cancer cohort from The Ohio State University Wexner Medical Center, consisting of 60 hematoxylin and eosin-stained whole-slide images from independent patients with varying degrees of gland differentiation, as well as on public benchmarks including the Gland Segmentation dataset (derived from stage T3–T4 colorectal adenocarcinomas), TCGA-COAD, TCGA-READ, and SPIDER.

**Results:** The proposed framework achieved strong performance on the institutional dataset despite limited annotations. On the Gland Segmentation dataset, it demonstrated competitive performance compared to both weakly and fully supervised approaches, achieving a mean Intersection over Union of 80.10% ± 1.52 and a mean Dice coefficient of 89.10% ± 2.10. Moreover, cross-cohort evaluations showed robust generalization on TCGA-COAD and TCGA-READ without requiring additional annotations, while reduced performance on SPIDER reflected pronounced domain shift.

**Conclusions:** Our framework provides an annotation-efficient and generalizable paradigm for accurate gland segmentation in colorectal histopathology, offering a practical pathway toward significantly reducing annotation burdens while preserving high segmentation fidelity.

# 1. Introduction

Colorectal cancer (CRC) is a leading cause of global cancer mortality, with epidemiological projections indicating a 60% increase in incidence by 2030, resulting in approximately 2.2 million new cases and 1.1 million fatalities annually.[1] Histopathological assessment of tissue architecture using hematoxylin and eosin (H&E)–stained sections is the gold standard for evaluating glandular structures in various types of adenocarcinomas, including CRC,[2] breast cancer,[3] prostate cancer,[4] and endometrial adenocarcinoma.[5] Histopathologic grading relies heavily on the degree of gland formation: well- and moderately differentiated tumors are classified as low grade, retaining largely intact glandular architecture, whereas poorly differentiated tumors display markedly complex, abortive, or absent glandular formation and are associated with worse survival outcomes.[6-8] Consequently, accurate gland segmentation in digitized whole-slide images (WSIs) is crucial for quantifying and characterizing glandular morphology, which directly informs tumor grading and risk stratification of colorectal and other types of adenocarcinomas.[3-5]

Fully supervised deep learning methods have set the benchmark for gland segmentation in histopathological images. Early work by DCAN established a multi-task learning framework that simultaneously segments glands and their contours to delineate benign, malignant, and closely apposed glands, and subsequent advances have incorporated domain-specific inductive biases,[9] such as Gabor-based encoders and topology-aware networks.[10],[11] To address residual segmentation errors, especially in apposed glands, Xie *et al*.[12] introduced a Deep Segmentation-Emendation model, which employs a dedicated emendation network to predict and correct inconsistencies in initial segmentation masks. While fully supervised methods demonstrate impressive performance, their effectiveness is intrinsically contingent upon the availability of large-scale, pixel-level annotated datasets—a major bottleneck in clinical practice due to the significant time and expertise required from pathologists.

This annotation burden has spurred interest in weakly supervised semantic segmentation (WSSS), which substantially reduces the demand for dense labels by leveraging weaker forms of supervision such as image-level labels or sparse annotations, thereby reducing annotation time by approximately sixty-fold.[13,14] The predominant WSSS approach uses classification networks to generate class activation maps (CAMs) as initial pseudo-labels for training segmentation models.[15-24] However, CAMs have inherent limitations; they tend to activate only the most discriminative regions of an object, resulting in pseudo-masks with ambiguous boundaries, noise, discontinuity, and structural fragmentation.[19] Various CAM refinement techniques have been developed to address these limitations, including SEAM, which enforces spatial consistency,[25] and AMR, which uses complementary activation branches to enhance under-activated regions.[26] Similarly, Kweon *et al.*[22] adopted an adversarial strategy, using an image reconstructor to force the classifier to generate more complete activation maps by minimizing inter-segment inferability. In medical imaging, domain-specific modifications such as C-CAM and MLPS have also been proposed.[13,14],[27]

Nevertheless, a significant challenge persists in learning from the noisy, CAM-generated pseudo-masks in the subsequent segmentation stage. Although recent frameworks such as ARML attempt to address both CAM refinement and noisy label learning in histopathology,[23] general WSSS methods still often underperform on gland segmentation due to the high morphological similarity between gland types and the critical need for precise instance boundaries. Therefore, there remains a significant methodological gap for a WSSS framework specifically designed to address the challenges of gland segmentation—namely, to generate high-quality, complete pseudo-masks from sparse annotations that can reliably guide the training of a dense segmentation model.

To bridge this gap, we introduced a novel weakly supervised teacher–student framework with progressive pseudo-mask refinement for multi-class gland segmentation in colorectal histopathology. The framework integrates an Exponential Moving Average (EMA)–stabilized teacher network with confidence-based filtering, curriculum-guided loss weighting, and adaptive pixel-wise fusion of sparse expert annotations to progressively discover and segment previously unannotated glandular structures.

Our contributions are threefold:

(1) We introduce a pixel-wise pseudo-label fusion strategy that preserves pathologist-provided sparse annotations while leveraging EMA-stabilized teacher predictions to supervise unlabeled regions during self-training.
(2) We propose a curriculum-driven refinement mechanism that combines cosine-decayed confidence thresholding with dynamic loss weighting, enabling progressive expansion of supervision from high-confidence gland regions to previously unannotated and ambiguous regions. This approach explicitly addresses annotation sparsity in the dense and morphologically complex setting of glandular histopathology.
(3) We perform a comprehensive, clinically grounded multi-cohort evaluation reflecting real-world variability. The framework is validated on (i) an institutional dataset with sparse annotations, (ii) the fully annotated public Gland Segmentation (GlaS) benchmark,[28] and (iii) three external cohorts—The Cancer Genome Atlas (TCGA) Colon Adenocarcinoma (COAD), Rectum Adenocarcinoma (READ), and SPIDER[29]—to assess cross-domain generalization. This multi-tiered evaluation demonstrates competitive performance relative to fully supervised methods and systematically characterizes robustness and failure modes under substantial domain shift, providing actionable insights for clinical translation.

# 2. Materials and methods

## 2.1. Study design and problem formulation

The proposed framework leverages the nnUNet backbone for robust semantic segmentation and comprises two identical networks: a student model ($\theta_S$) trained via gradient descent using a supervised segmentation loss and a consistency regularization term, and a teacher model ($\theta_T$) updated exclusively through an EMA of the student parameters, providing stable pseudo-labels that guide student learning. Formally, let $x \in \mathbb{R}^{H\times W\times 3}$denote an input image, and $y \in \{0,1,\dots,C\}^{H\times W}$ the corresponding pixel-level labels in the segmentation mask, where $C = 4$ represents the total number of classes (background stroma, benign glands, malignant glands, and poorly differentiated clusters/glands). The goal is to learn a function $f_\theta(x)$ that outputs pixel-wise class probabilities $p_\theta(x) \in [0,1]^{H\times W\times C}$ to segment both annotated and unannotated glandular structures at the pixel level.

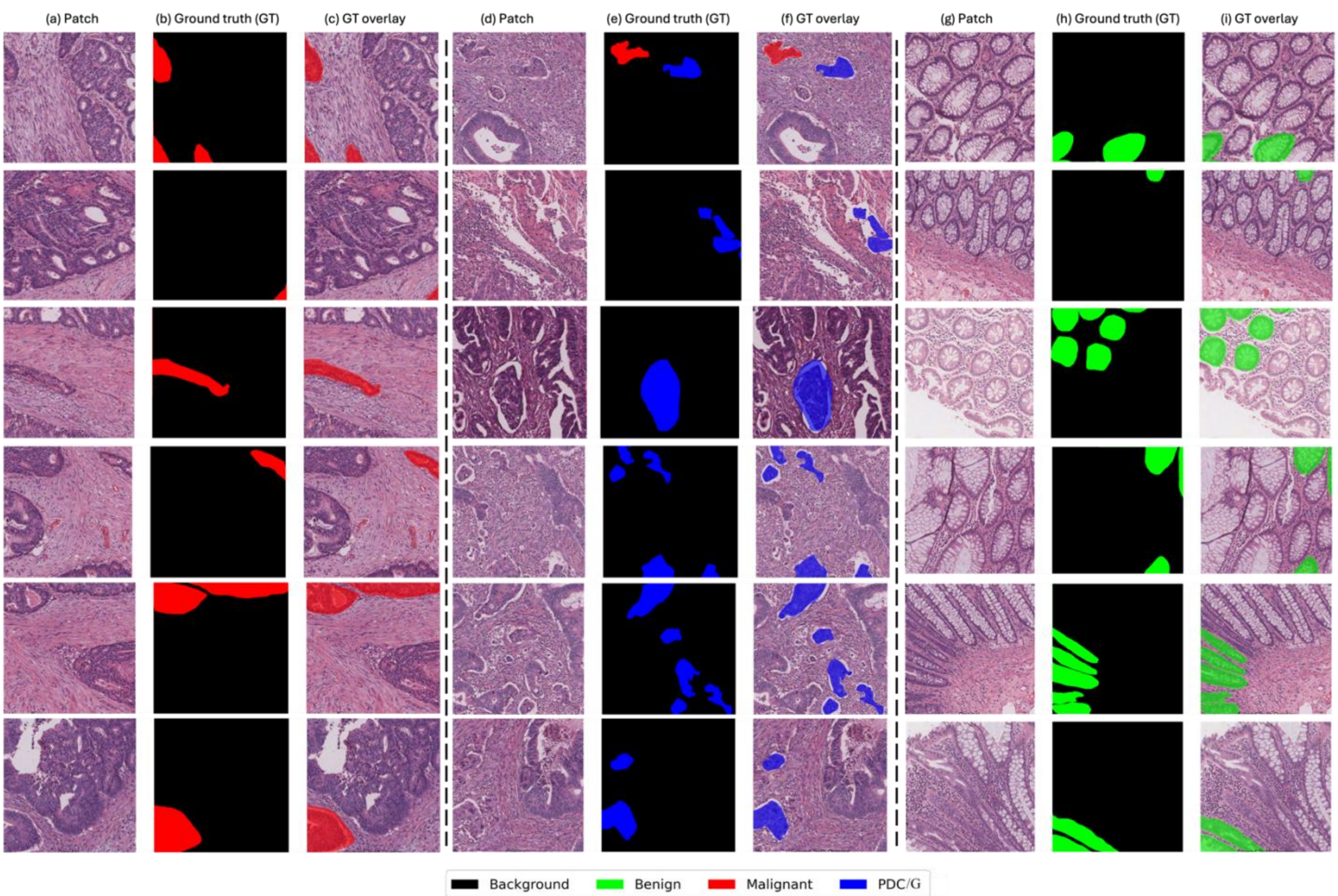

**Fig. 1. Representative samples from the in-house The Ohio State University Wexner Medical Center (OSUWMC) dataset, illustrating sparse annotations provided by pathologists for three key gland classes: benign glands, malignant glands, and poorly differentiated clusters/glands (PDC/G).** For each class, the column of triplets shows: (i) the original histopathology image, (ii) the corresponding sparse ground truth mask from the two experts, and (iii) an overlay of the annotations on the input image. Only a few sparsely annotated glandular structures are present within each patch, leaving substantial regions unlabeled, which significantly increases the difficulty of learning accurate segmentation models under weak supervision. Color coding for all segmentation masks is as follows: red represents malignant glands, green represents benign glands, blue indicates poorly differentiated clusters or glands (PDC/G), and black denotes stroma (best viewed in color).

### 2.2. Datasets

We conducted experiments on the in-house The Ohio State University Wexner Medical Center (OSUWMC) dataset containing limited pathologist annotations, as well as on the publicly available GlaS dataset with high-quality pixel-level annotations to demonstrate the broad applicability of the framework.[28] Additionally, three external publicly available CRC histopathology datasets, TCGA-COAD, TCGA-READ, and SPIDER[29], were used to qualitatively assess the generalizability of the proposed framework on external cohorts where ground-truth (GT) annotations are not available.

**a) OSUWMC in-house dataset:** We used an in-house CRC histology dataset collected at OSUWMC, consisting of 60 H&E-stained WSIs from independent patients with histologically confirmed colorectal adenocarcinoma. All WSIs were retrospectively acquired from surgical resection specimens, annotated by two pathology residents using sparse pixel-level labels, and scanned at 40× magnification. The dataset is WSIs only; no patient-level clinical or demographic metadata (e.g., age, sex, tumor stage, grade, or treatment history) were collected or were available, as this study focused exclusively on technical development of weakly supervised segmentation algorithms rather than clinical outcome prediction. The annotations include four tissue categories: benign glands, malignant glands (better-formed tumor glands with obvious lumina), poorly differentiated clusters/glands (encompassing tumor buds, poorly differentiated clusters, and poorly formed tumor glands with absent or minimal lumina), and background stroma. The cohort captures a broad range of glandular morphologies, including well-formed glands, irregular malignant glands, and poorly differentiated structures, reflecting real-world histopathologic variability. WSIs were scanned at 40× magnification, and 512 × 512-pixel patches were extracted at 5× magnification for model development. In total, 74,179 patches were generated and split into 63,191 training, 5,460 validation, and 5,528 test patches. Approximate class prevalence at the patch level was ~45% benign glands, ~35% malignant glands, ~15% background stroma, and ~5% PDC/G, with stratified sampling used to preserve class proportions across splits. Figure 1 shows representative samples from our in-house dataset, which contains sparse annotations for background stroma, benign glands, malignant glands, and poorly differentiated clusters/glands. Notably, most patches contained both annotated and unannotated glands, posing a significant challenge for accurate segmentation under weak supervision.

**b) GlaS Dataset:** We subsequently conducted experiments using the GlaS dataset,[28] a publicly available histological image collection released as part of the MICCAI 2015 Gland Segmentation Challenge. The dataset comprises 165 H&E-stained images extracted from 16 colorectal tissue sections, each obtained from a different patient diagnosed with stage T3 or T4 colorectal adenocarcinoma. All cases correspond to advanced-stage disease, and no earlier-stage tumors are included in the cohort.[28] Per the official GlaS challenge protocol, patient-level demographic information (e.g., age, sex, exact TNM substage) is not provided with the dataset and is not required for the benchmark segmentation task, which is defined strictly at the image and pixel level. The images were scanned at 20× magnification with a native spatial resolution of 0.465 μm/pixel, and most have an original size of 775 × 522 pixels. Each image is accompanied by instance-level segmentation ground truth, providing precise delineation of glandular boundaries. Within each image, both benign and malignant glandular structures are present, reflecting the heterogeneous histologic architecture typical of advanced colorectal adenocarcinoma. The dataset is divided into 85 training images (37 benign and 48 malignant) and 80 test images (37 benign and 43 malignant). This benign/malignant distribution at the image level is reported in accordance with established benchmark practice and provides sufficient characterization for the segmentation task.[28] To ensure consistency with prior work and standardize input resolution, all images were resized to 512 × 512 pixels. For model development, the training set was further partitioned into 70 images for training (~82.4% of the training set) and 15 for validation (~17.6% of the training set) using a stratified sampling strategy to preserve the benign–malignant class balance, while the 80 test images were used exclusively for final performance evaluation. Because pathological stage is fixed (T3–T4) across the dataset, no stage-based stratification was required during training or evaluation. A key challenge posed by GlaS is the substantial inter-subject variability in staining characteristics and tissue morphology, arising from differences in laboratory processing, which makes the dataset a rigorous benchmark for gland segmentation algorithms.

### 2.3. Two-phase training protocol

Figure 2 illustrates a schematic overview of the proposed framework, which comprises two phases: a supervised warm-up phase and a teacher–student co-training phase.

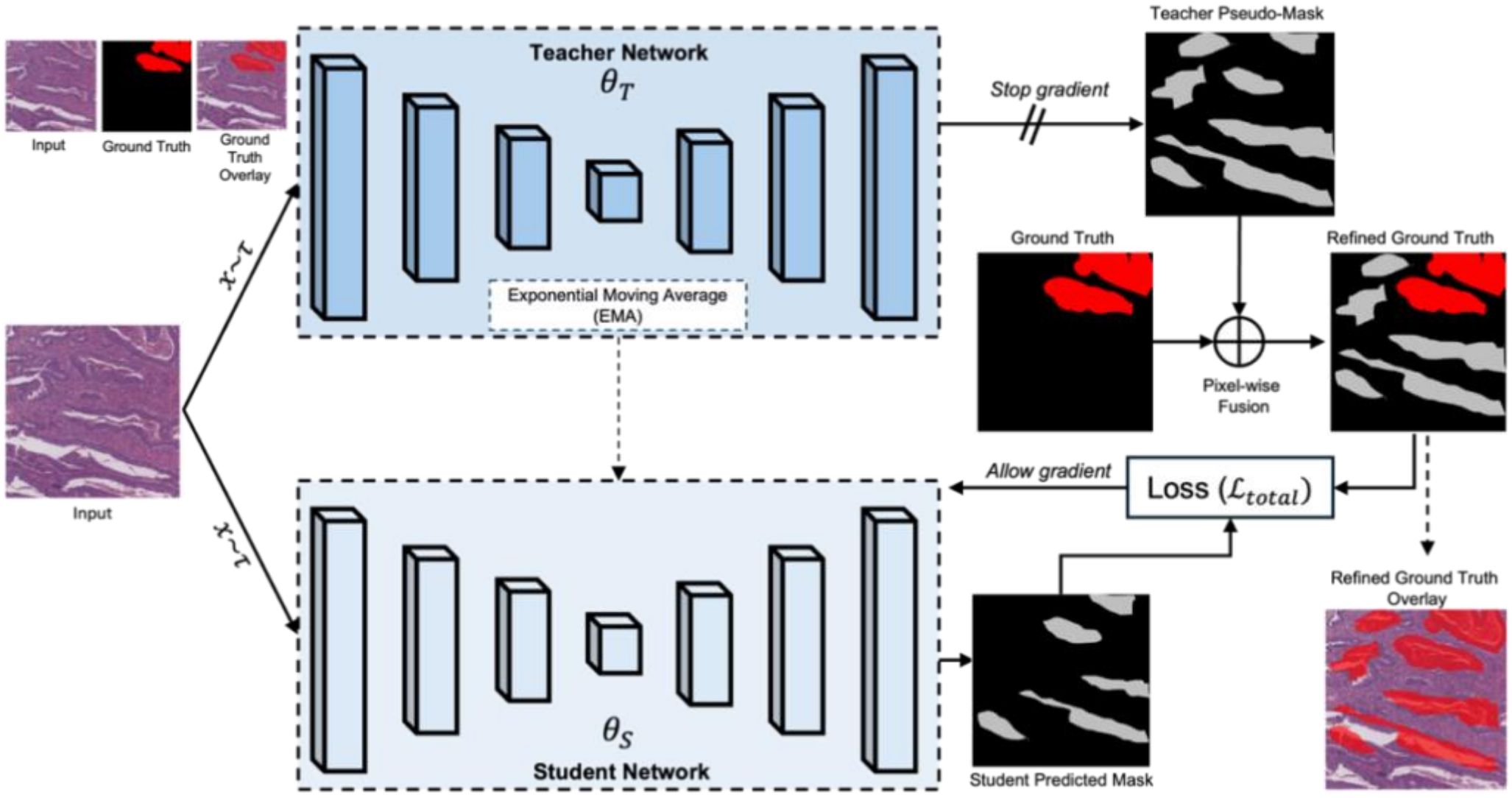

**Fig. 2.** Schematic overview of the proposed teacher–student self-training framework for weakly supervised multi-class gland segmentation. The teacher network ($\theta_T$), stabilized by an Exponential Moving Average (EMA), generates initial pseudo-masks. The initial pseudo-masks are then refined via confidence-based filtering and adaptively fused with sparse ground-truth annotations to produce high-quality supervision for training the student network ($\theta_S$). The student's parameters are then used to update the teacher via EMA. This iterative process, governed by a total loss ($\mathcal{L}_{total}$), enables progressive discovery and segmentation of unannotated glandular structures (best viewed in color).

**Phase 1: Supervised warm-up**

During the warm-up phase, the teacher network remains inactive, and the student network is trained solely on the available sparse annotations. This strategy ensures the student learns robust and meaningful representations that are essential for subsequent pseudo-label generation. The student network is optimized using a supervised loss ($\mathcal{L}_{\text{supervised}}$), defined as a weighted combination of Dice loss ($\mathcal{L}_{dice}$) and categorical cross-entropy loss ($\mathcal{L}_{cce}$) as follows:

$$\mathcal{L}_{\text{supervised}} = \mathcal{L}_{dice} + \mathcal{L}_{cce} \tag{1}$$

Here, $\mathcal{L}_{dice}$ maximizes the overlap between predicted and ground truth masks, while $\mathcal{L}_{cce}$ evaluates the pixel-wise classification accuracy across the $\mathrm{C} = 4$ classes. Formally, these losses are defined as follows.

$$\mathcal{L}_{dice} = 1 - \frac{2\sum_i y_{i,c} \cdot \hat{y}_{i,c}}{\sum_i y_{i,c} + \sum_i \hat{y}_{i,c}} \tag{2}$$

$$\mathcal{L}_{cce} = -\frac{1}{N}\sum_{i=1}^{N}\sum_{c=1}^{\mathrm{C}=4} y_{i,c} \log\left(\hat{y}_{i,c}\right) \tag{3}$$

where $N$ denotes the total number of pixels, $y_{i,c} \in \{0,1\}$ indicates whether pixel $i$ belongs to class $c$, and $\hat{y}_{i,c}$ represent the predicted probability of class $c$ at pixel $i$. The warm-up phase typically spans 20% to 25% of the total epochs, providing a stable initialization for the teacher network.

**Phase 2: Teacher–Student co-training**

Upon completion of the warm-up phase, the teacher is initialized with the student's parameters, i.e., $\theta_T \leftarrow \theta_S$. Subsequently, the student network ($\theta_S$) is optimized via gradient descent, while the teacher network ($\theta_T$) is updated using an EMA of the student's weights. Formally, the teacher parameters are updated as follows:

$$\theta_T \leftarrow \beta\theta_T + (1-\beta)\,\theta_S \tag{4}$$

where the EMA decay coefficient $\beta$ is set to 0.999 to ensure temporally smooth teacher updates and to suppress short-term fluctuations in the student model. A high decay value is particularly important in weakly supervised dense segmentation settings, as it stabilizes pseudo-label generation and mitigates confirmation bias arising from noisy early predictions. This choice is consistent with prior teacher–student and Mean Teacher frameworks, which commonly adopt decay values in the range of 0.99–0.999 for segmentation tasks.

This update strategy yields temporally smooth teacher predictions, enhancing pseudo-label stability and mitigating confirmation bias induced by noisy CAMs. During this phase, the student is trained using a hybrid loss that integrates supervised learning on labeled data with consistency regularization provided by the teacher as follows.

$$\mathcal{L}_{total} = \alpha(t)\mathcal{L}_{supervised} + \big(1-\alpha(t)\big)\,\mathcal{L}_{consistancy} \tag{5}$$

Here, $\alpha(t)$ is a dynamic, epoch-dependent weighting factor that governs the trade-off between the supervised and consistency losses. We employ a cosine-decaying schedule for $\alpha(t)$ to gradually shift emphasis from GT supervision to teacher-guided consistency. After warm-up, $\alpha(t)$ is initialized at 0.9, placing 90% reliance on supervised loss, and decays to 0.01 by the end of training, progressively increasing reliance on teacher-generated pseudo-labels. The smooth cosine decay prevents abrupt transitions, reduces early over-reliance on noisy pseudo-labels, and enables stable late-stage refinement.

**Teacher-generated pseudo-mask:** The consistency term in Eq. (5) encourages the student to align with the teacher's segmentation predictions on both labeled and unlabeled pixels. To ensure the reliability of the teacher-generated pseudo-labels, we employ a confidence-based filtering mechanism that suppresses low-confidence or ambiguous pseudo-labels, particularly during the early phase of training. Formally, the confidence mask is defined as follows:

$$m(x) = 1\big[max\big(\sigma(f_{\theta_T}(x))\big) > \tau_{confidence}(t)\big] \tag{6}$$

where $\sigma(.)$ is the softmax, $1[.]$ is the indicator function, and $\tau_{confidence}(t)$ is a cosine-decaying threshold that monotonically decreases from 0.95 to 0.25 over the course of training. The high initial threshold restricts supervision to only the most confident teacher predictions when the teacher model is still stabilizing, while the gradual relaxation allows progressively more ambiguous regions—such as gland boundaries and poorly differentiated structures—to be incorporated as training proceeds. This curriculum-guided design enables stable expansion of pseudo-label coverage while minimizing noise propagation.

These bounds are empirically selected to incorporate a curriculum learning strategy that emphasizes high-confidence teacher pixel-level supervision in early training and gradually incorporates other teacher-generated pseudo-labels for unlabeled regions as the teacher stabilizes. To maximally leverage sparse annotations, the teacher-generated pseudo-labels are fused with GT labels using a pixel-wise integration strategy defined as follows:

$$m(x) = \begin{cases} GT(m(x)), & if \quad GT(x) > 0 \\ m(x), & otherwise. \end{cases} \tag{7}$$

This formulation ensures that pathologist-provided annotations are preserved exactly in labeled regions, while teacher-generated pixel-level pseudo-masks supervise the unlabeled regions. This fusion strategy is employed only after the teacher model has reached sufficient stability. The consistency loss is defined as follows:

$$\mathcal{L}_{consistency} = \left\|\sigma\left(f_{\theta_S}(x)\right) - m(x)\right\|^2 \tag{8}$$

where $\sigma(.)$ denotes the softmax function, converting the student network output logits into per-pixel class probabilities. We employ logit-level mean squared error for consistency regularization, which empirically stabilizes training and reduces sensitivity to early-stage noise in the pseudo-labels.

### 2.4. Baselines

To benchmark the efficacy of our proposed framework, we evaluated its performance against a comprehensive set of existing methods, including thirteen WSSS and eleven fully supervised segmentation approaches.[30-32] The WSSS baselines include SEAM,[25] ReCAM,[21] AMR,[26] MLPS,[14] OEEM,[33] AME-CAM,[34] HAMIL,[35] CBFNet MPFP,[31],[36] Adv-CAM,[37] SC-CAM, and MAA.[15],[30,31] The fully supervised baselines consist of widely used architectures: UNet,[38] Seg-Net,[39] MedT,[40] TransUNet,[41] Attention Unet,[42] UNet++,[43] KiU-Net,[44] ResUNet++,[45] DA-TransUNet,[41] TransAttUNet,[46] and EWASwin UNet.[32] To ensure a fair comparison, we adhered to the experimental protocols and key hyperparameters (e.g., patch size) specified in the respective original baseline publications.[30-32]

### 2.5. Implementation details

All experiments were conducted using PyTorch 1.13.1 with CUDA 11.7 on Python 3.10. Training was performed on NVIDIA A100 GPUs. To ensure reproducibility, the random seed was fixed at 42 across all libraries (Python, NumPy, PyTorch, and CUDA), and deterministic algorithms were enforced. However, to quantify statistical variability, we performed five independent training runs with different random seeds for all experiments and report the mean ± standard deviation across these runs. The models were trained using the AdamW optimizer with an initial learning rate of 0.01 and a weight decay of 0.001.[47] A cosine annealing schedule was employed to decay the learning rate to a minimum of 0.00001. We used a batch size of 16 and an input patch resolution of $512 \times 512$ pixels. To stabilize training, gradient clipping was applied with a maximum norm of 1.0. To enhance generalization, we utilized a comprehensive data augmentation strategy, including random discrete rotations ($0°, 90°, 180°, 270°$), horizontal flipping ($P = 0.5$), hue–saturation–value jittering, Gaussian noise, and Gaussian blur, followed by standard ImageNet normalization.[48] The maximum training duration was set to 250 epochs, with an early stopping mechanism triggered to prevent overfitting if validation performance did not improve for 50 consecutive epochs.

### 2.6. Evaluation metrics

We employed two widely adopted metrics in gland segmentation[49]: mean Intersection over Union (mIoU) and mean Dice coefficient (mDice). Both metrics are derived from pixel-level classification outcomes, where each pixel is categorized as true positive (TP), false positive (FP), or false negative (FN) with respect to the GT annotation. The mIoU measures the overlap between the predicted and GT gland regions and is defined as:

$$\text{mIoU} = \frac{TP}{TP + FP + FN},$$

The mDice evaluates the similarity between the predicted mask and the ground truth and is formulated as:

$$\text{mDice} = \frac{2 \times TP}{2 \times TP + FP + FN}.$$

Both metrics are normalized to the range $[0, 1]$, where a value of 1 indicates perfect alignment between the prediction and the ground truth, and 0 implies no overlap. Higher scores correspond to superior segmentation accuracy and better boundary delineation.

## 3. Results

To validate the efficacy of our proposed framework, we assessed the proposed framework using the public GlaS dataset with dense annotations and an in-house OSUWMC cohort to evaluate performance under sparse-label conditions. Generalization beyond the training domain was examined by applying the model trained on the OSUWMC cohort to the TCGA-COAD, TCGA-READ, and SPIDER datasets. As GT annotations are unavailable for these external cohorts, evaluation was limited to qualitative analysis. The proposed framework was compared against a broad range of state-of-the-art approaches, including weakly supervised methods (summarized in Table 1) and fully supervised architectures (summarized in Table 2).

### 3.1. Performance against weakly supervised methods

As Table 1 summarizes, our framework achieves competitive state-of-the-art performance on the GlaS benchmark,[30,31] achieving an mIoU of 80.10% and an mDice of 89.10%, while our mIoU is slightly below that of the leading MAA method. The proposed framework demonstrates markedly superior training stability, evidenced by a lower variance (±1.52 mIoU, ±2.10 mDice) compared to MAA (±2.26 mIoU, ±3.31 mDice). The high consistency and lower variance are critical prerequisites for clinical translation and underscore the robustness of our pseudo-label refinement strategy.

### 3.2. Performance against fully supervised methods

As Table 2 summarizes, our framework achieves competitive performance compared to fully supervised state-of-the-art methods on GlaS.[32] Specifically, our framework attains 0.801 mIoU and 0.891 mDice, which are on par with the top-performing supervised baseline, EWASwin UNet (0.815 mIoU).[32] Moreover, our framework surpasses traditional architectures such as UNet++ and ResUNet++ (mIoU ~0.70–0.74) as well as other advanced models such as TransAttUNet (0.777 mIoU).[43],[45],[46] These quantitative findings are corroborated by the qualitative comparisons shown in Figure 3 (with additional examples in Figure 6), which demonstrate the model's ability to generate precise segmentation masks [32]. Notably, these results underscore the capability of our weakly supervised framework to effectively leverage sparse annotations to obtain performance competitive with leading fully supervised methods.

**Table 1.** Comparison with weakly supervised gland segmentation methods on the GlaS dataset

| **Method** | **Year** | **mIoU (%)** | **mDice (%)** |
|---|---|---|---|
| SEAM[25] | 2020 | 71.36±0.49 | 79.59±4.88 |
| ReCAM[21] | 2022 | 56.31±2.53 | - |
| AMR[26] | 2022 | 72.83±0.37 | - |
| MLPS[14] | 2022 | 73.60±0.16 | - |
| OEEM[33] | 2022 | 76.48±0.10 | 83.40±5.36 |
| AME-CAM[34] | 2023 | 74.09±0.13 | - |
| HAMIL[35] | 2023 | 77.37±0.73 | - |
| CBFNet[36] | 2024 | 76.30±0.26 | - |
| MPFP[31] | 2025 | 80.44±0.05 | - |
| Adv-CAM[37] | 2021 | 68.54±3.36 | 81.33±5.26 |
| SC-CAM[15] | 2020 | 71.52±3.50 | 83.40±5.36 |
| MAA[30] | 2025 | 81.99±2.26 | 90.10±3.31 |
| Ours | - | 80.10±1.52 | 89.10±2.10 |

Metrics are mean Intersection over Union (mIoU) and mean Dice coefficient (mDice), reported as mean ± standard deviation across five independent training runs with different random seeds. Our framework achieves competitive performance compared with current state-of-the-art approaches.

**Table 2.** Comparison with fully supervised gland segmentation methods on the GlaS dataset

| Method | Year | mIoU (%) | mDice (%) |
|---|---|---|---|
| UNet[38] | 2015 | 64.8 | 77.6 |
| Seg-Net[39] | 2017 | 66.0 | 78.6 |
| MedT[40] | 2021 | 69.6 | 81.0 |
| TransUNet[41] | 2021 | 70.1 | 81.5 |
| AttentionUNet[42] | 2018 | 70.1 | 81.6 |
| UNet++[43] | 2018 | 70.2 | 81.9 |
| KiU-Net[44] | 2020 | 72.8 | 83.3 |
| ResUNet++[45] | 2019 | 73.8 | 84.1 |
| DA-TransUNet[41] | 2024 | 75.6 | 85.3 |
| TransAttUNet[46] | 2023 | 77.7 | 86.7 |

| EWASwin UNet[32] | 2025 | 81.5 | 89.4 |
|---|---|---|---|
| Ours | - | 80.1 | 89.1 |

Our framework achieves competitive performance among current state-of-the-art approaches, including classical CNN-based models, transformer-enhanced architectures, and recent hybrid methods, as measured by mean Intersection over Union (mIoU) and mean Dice coefficient (mDice). Although designed for limited-annotation scenarios, the performance of our framework improves with the availability of higher-quality annotations.

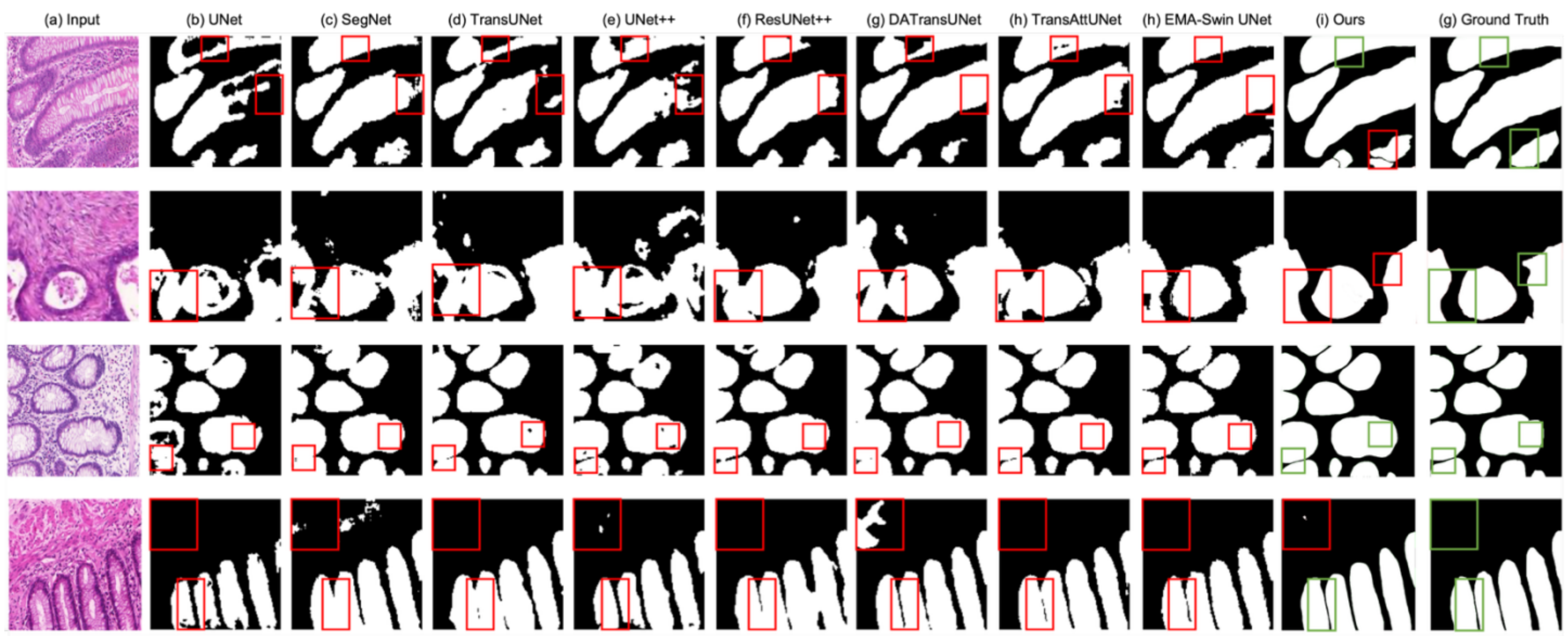

**Fig. 3.** Results on the GlaS dataset [32]. In each row, the leftmost image shows the original H&E patch, followed by segmentation results from baseline methods, with the rightmost column displaying the ground truth annotations (best viewed in color).

### 3.3. Results on the OSUWMC dataset and out-of-domain generalization to TCGA-COAD, TCGA-READ, and SPIDER

Figure 4 illustrates the qualitative performance of our framework on the in-house OSUWMC dataset. These visualizations demonstrate how the stabilized teacher network effectively guides the student model via pseudo-masks, enabling the discovery and precise segmentation of unannotated gland structures using only limited supervision. To evaluate robustness and clinical transferability, we performed whole-slide inference on three external cohorts, including TCGA-COAD, TCGA-READ, and SPIDER. Despite significant inter-institutional variations in staining protocols and scanner characteristics, our model maintained consistent qualitative performance (see Fig. 5), successfully identifying benign glands, malignant glands, and poorly differentiated clusters/glands on TCGA-COAD and TCGA-READ. In contrast, on the SPIDER dataset, we observed notable qualitative performance degradation, characterized by fragmented gland boundaries, increased FPs in stromal regions, and reduced sensitivity to poorly differentiated glandular structures. Quantitative evaluation was not performed, as pixel-level gland annotations are not available for these datasets. The observed performance degradation is therefore reported qualitatively and is attributed to severe domain shift, including lower image quality, pronounced staining heterogeneity, and higher morphological variability inherent to that specific cohort, highlighting the challenges of cross-domain generalization in histopathology.[50,51]

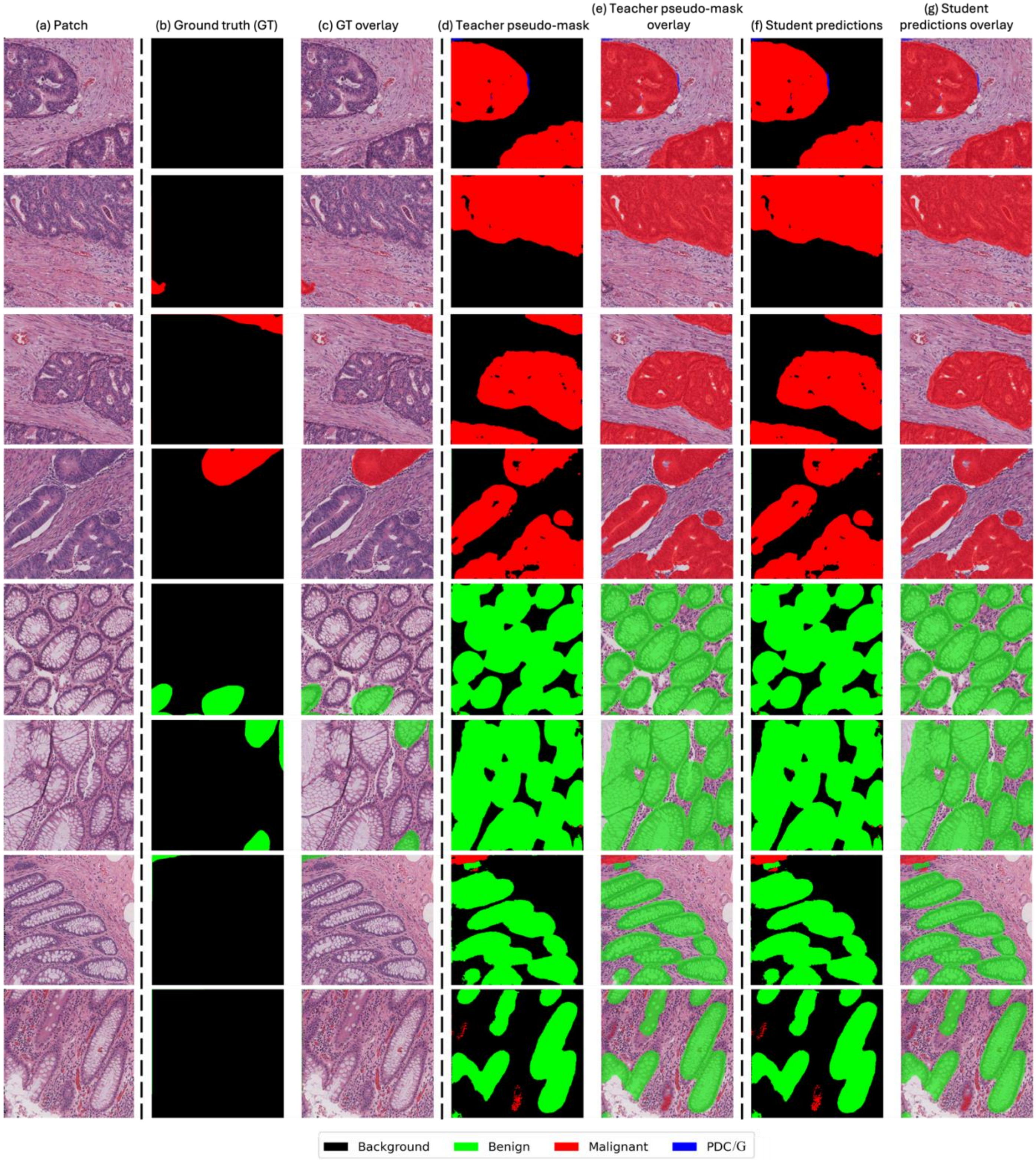

**Fig. 4.** Qualitative segmentation results on the in-house Ohio State University Wexner Medical Center (OSUWMC) dataset. For each representative sample: (a) input H&E image; (b) sparse ground-truth annotations provided by two pathologists; (c) annotation overlay on the input image; (d) pseudo-mask generated by the teacher model; (e) teacher pseudo-mask overlay; (f) final prediction from the student model; and (g) student prediction overlay. Color coding: red = malignant glands, green = benign glands, blue = poorly differentiated clusters/glands (PDC/G), black = background stroma. This visualization highlights the framework's ability to perform gland segmentation under sparse or limited-annotation supervision (best viewed in color).

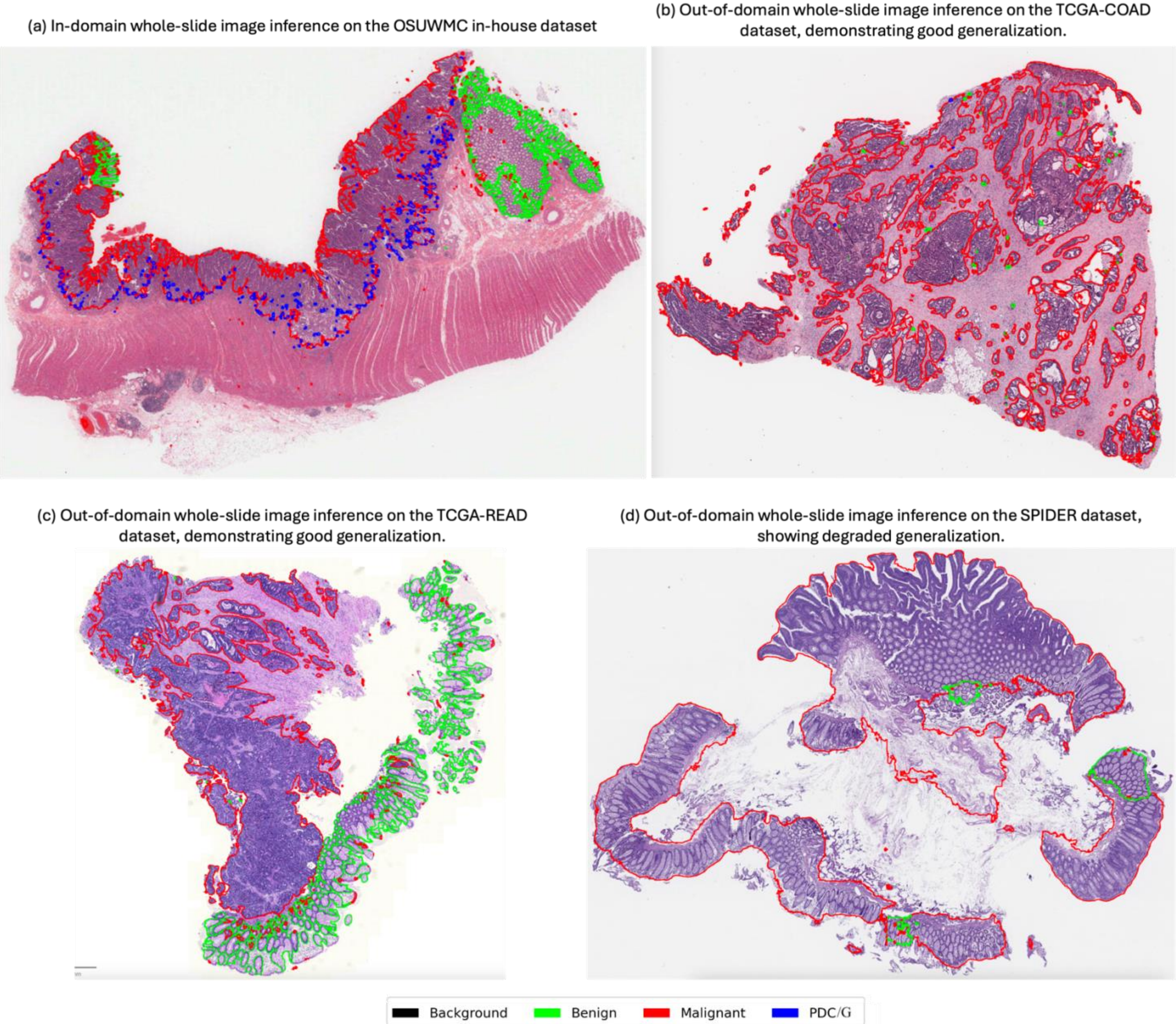

**Fig. 5.** Whole-slide–level qualitative assessment across (a) in-house OSUWMC, (b) TCGA-COAD (The Cancer Genome Atlas Colon Adenocarcinoma), (c) TCGA-READ (Rectal Adenocarcinoma), and (d) SPIDER, illustrating cross-domain generalization. Color coding: red = malignant glands, green = benign glands, blue = poorly differentiated clusters/glands (PDC/G), black = background stroma. Our framework performs robustly on OSUWMC with limited ground-truth annotation and on TCGA-COAD and TCGA-READ without any ground-truth annotation. Performance degradation on SPIDER—also evaluated without any annotation—is attributed to lower image quality, staining heterogeneity, and substantial domain shift (best viewed in color).

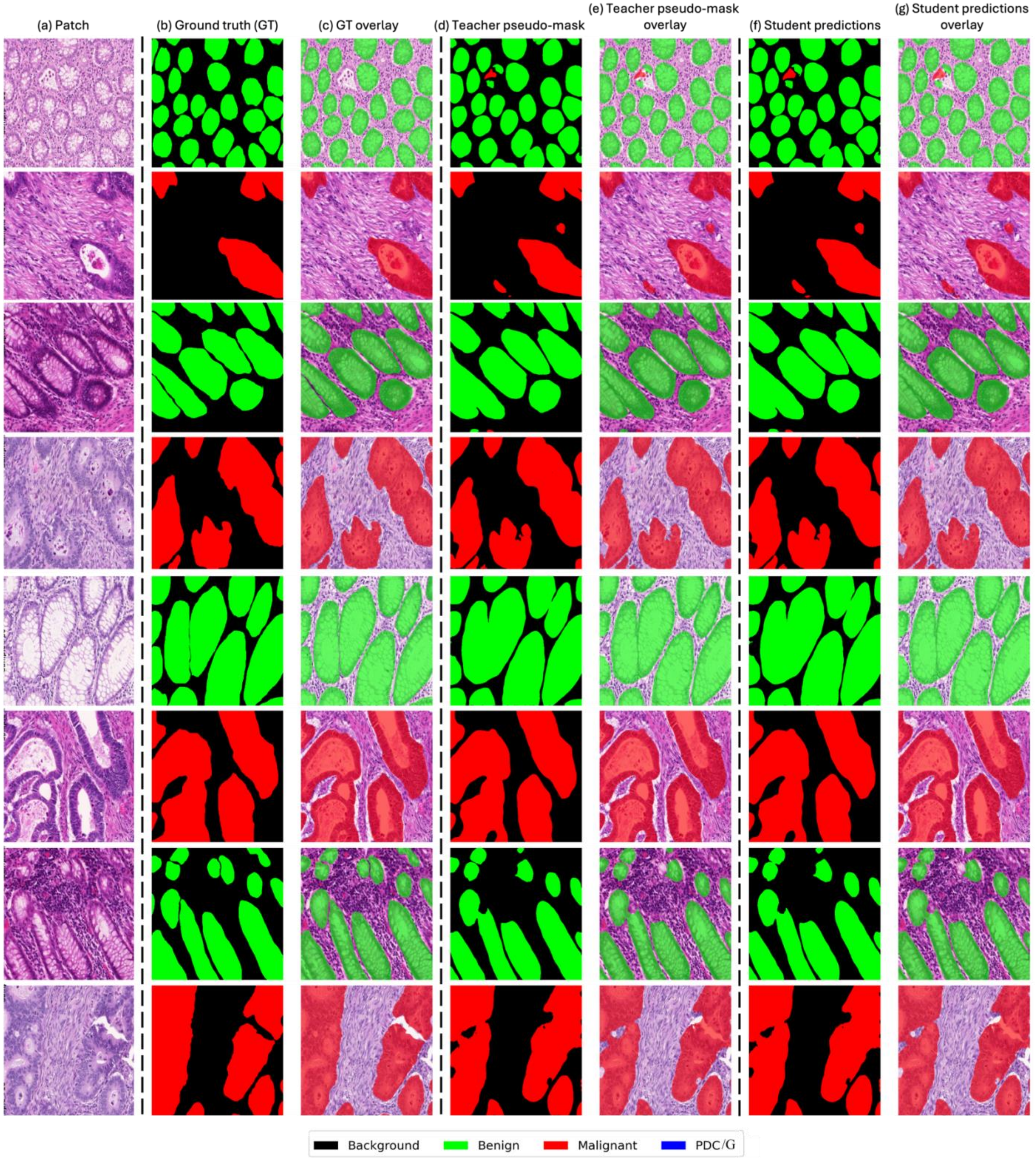

**Fig. 6.** Qualitative results on the GlaS test set. For each sample: (a) original H&E image; (b) dense ground truth mask; (c) overlay of ground truth on the input; (d) pseudo-mask predicted by the teacher model; (e) overlay of the teacher's prediction; (f) final segmentation mask predicted by the student model; and (g) overlay of the student's prediction. Color coding: red = malignant glands, green = benign glands, black = background stroma (best viewed in color).

# 4. Discussion

We developed and validated a novel weakly supervised teacher–student framework for multi-class gland segmentation in CRC histopathology. The core of our approach lies in an EMA-stabilized teacher network, which employs confidence-based filtering and an adaptive fusion strategy to iteratively refine pseudo-masks, thereby guiding the student network with increasingly reliable supervision. Notably, the competitive results on the GlaS benchmark indicate that access to high-quality annotations allows our framework to further narrow the performance gap with fully supervised methods. In clinical settings, where dense, pixel-level annotation remains a major bottleneck, our framework offers practical benefits by substantially reducing annotation requirements while maintaining strong segmentation performance. Furthermore, its ability to generalize to TCGA-COAD and TCGA-READ without additional fine-tuning underscores its potential for multi-center application, where variations in staining and scanning protocols are common.

However, the performance drop on SPIDER highlights the well-known challenge of domain generalization in computational pathology.[50,51] While our method generalizes well to TCGA-COAD and TCGA-READ domains (similar domains), SPIDER represents a severe domain shift that likely requires explicit domain adaptation techniques. Future work will focus on incorporating advanced domain adaptation strategies to improve broader cross-institutional generalization.[50,51] Additionally, we plan to extend the framework to other adenocarcinoma types, such as prostate, breast, and lung cancers, where glandular segmentation is equally critical for diagnosis and grading. By further reducing reliance on manual annotations while maintaining high segmentation fidelity, our framework offers a scalable and practical pathway toward wider adoption of computational pathology tools in clinical workflows.

# 5. Limitations

Despite the promising performance of the proposed framework, several limitations merit consideration. First, the OSUWMC dataset lacks patient-level clinical metadata, precluding clinicopathologic correlation analyses. While this does not affect the technical validity of pixel-wise segmentation evaluation, it limits assessment of downstream prognostic or clinical utility. Second, although results are reported as mean ± standard deviation across independent runs, additional statistical measures, e.g., confidence intervals, will be investigated in future work. Third, the performance degradation observed on the SPIDER dataset highlights the impact of severe domain shift; addressing this limitation will likely require explicit domain adaptation or stain normalization strategies. Finally, while the proposed framework substantially reduces annotation burden, it still relies on limited sparse expert annotations. Achieving fully annotation-free segmentation remains an open and important direction for future research.

# 6. Conclusions

In this study, we introduced a novel weakly supervised teacher–student framework for multi-class gland segmentation in CRC histopathology, specifically designed to overcome the critical bottleneck and demand for extensive pixel-level annotations. By leveraging an EMA-stabilized teacher network, our framework efficiently utilizes sparse annotations, progressively refining pseudo-labels through confidence-based filtering and adaptive GT fusion. Comprehensive evaluation demonstrates that our framework achieved performance competitive with state-of-the-art methods in both weakly and fully supervised settings. Furthermore, the model exhibits strong in-house performance and robust generalization to external cohorts, including TCGA-COAD and TCGA-READ. While performance limitations on SPIDER highlight challenges under extreme domain shift, overall this work establishes an annotation-efficient paradigm that directly addresses a fundamental impediment in computational pathology. By substantially reducing reliance on costly manual curation while maintaining high segmentation fidelity, the proposed framework offers a practical, translatable solution to accelerate the adoption of automated diagnostic tools in clinical workflows.

**Acknowledgments**

**Conflict of interest**

The authors declare no competing interests.

**Funding**

This project was supported by R01 CA276301 (PIs: Niazi, Chen) from the National Cancer Institute. The project was also supported by The Ohio State University Comprehensive Cancer Center, Pelotonia Research Funds, and the Department of Pathology. The content is solely the responsibility of the authors and does not necessarily represent the official views of the National Institutes of Health or the National Cancer Institute.

**Author contributions**

Leadership, experimental design, data analysis, figure and table preparation, manuscript drafting (HK), funding acquisition, and writing—review and editing (WC, MKKN). All authors have approved the final version and publication of the manuscript.

**Ethical statement**

The use of the in-house OSUWMC dataset was approved by the Institutional Review Board of The Ohio State University Wexner Medical Center (IRB No. 2018C0098). Written informed consent was obtained from all patients or was waived by the IRB due to the retrospective nature of the study. Public datasets (TCGA and SPIDER) were used in compliance with their respective data usage agreements and ethical guidelines and do not require additional institutional approval. All procedures were performed in accordance with the ethical standards of the Declaration of Helsinki (as revised in 2024).

**Data sharing statement**

The in-house OSUWMC dataset used in this study is available upon reasonable request by contacting the corresponding author, Hikmat Khan, at Hikmat.Khan@osumc.edu | Hikmat.khan179@gmail.com. All code was implemented in Python using PyTorch as the primary deep-learning library. The complete pipeline for processing WSIs, as well as training and evaluating the deep-learning models, will be available at: https://github.com/hikmatkhan/gland-segmentation-teacher-student.

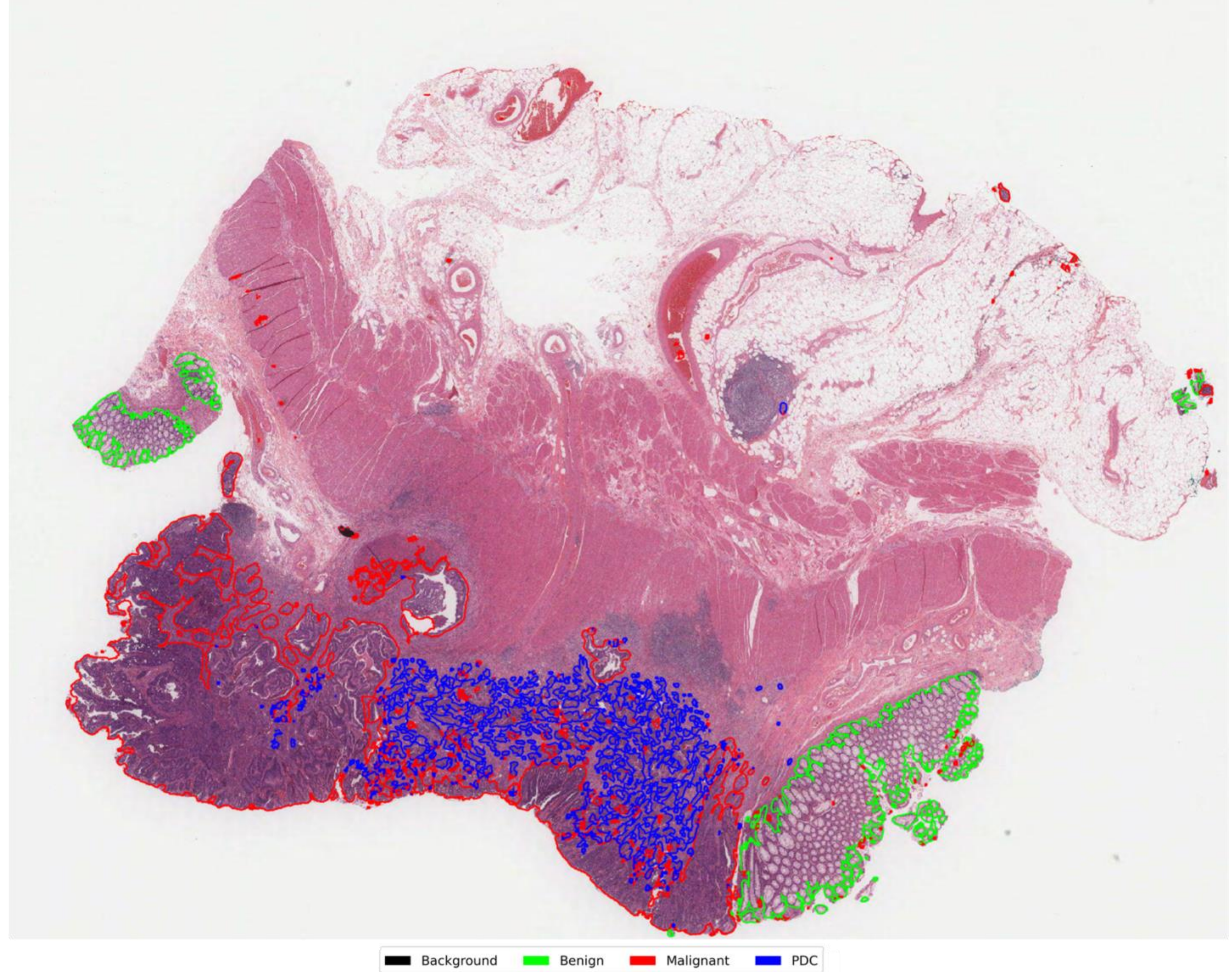

**Fig. 7.** Visualization of gland segmentation annotations generated by the proposed framework. A representative hematoxylin and eosin (H&E)–stained colorectal histology patch with segmentation annotations produced by the proposed weakly supervised teacher–student framework. The color-coded contours indicate different glandular categories predicted by the model: red denotes malignant glands, green denotes benign glands, and blue indicates poorly differentiated clusters/glands (PDC/G). Regions without colored annotations correspond to background stroma. This example illustrates the framework's ability to identify and delineate multiple glandular structures across the tissue section under weak supervision. Color coding: red = malignant glands, green = benign glands, black = background stroma (best viewed in color).

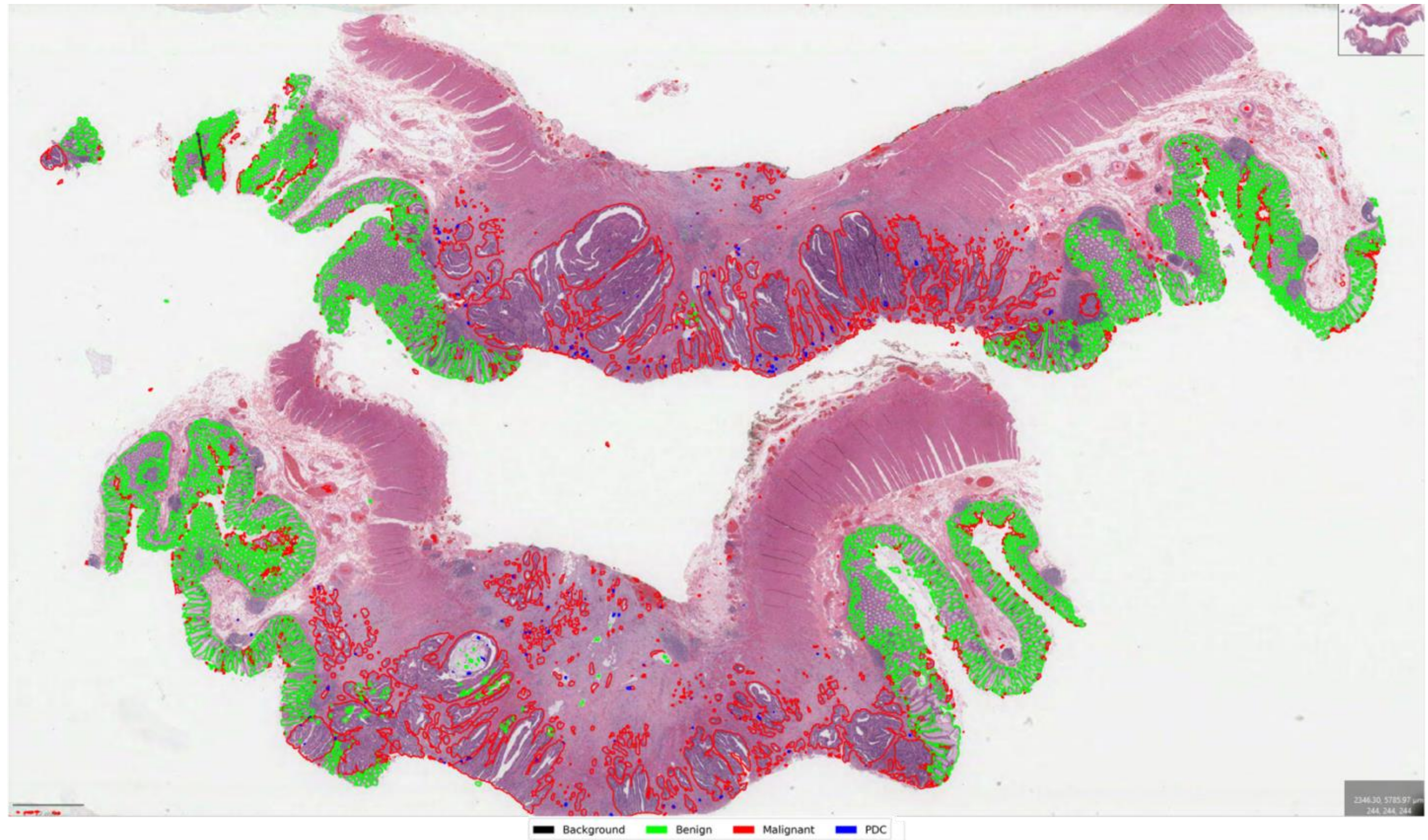

**Fig. 8.** Visualization of gland segmentation annotations generated by the proposed framework. A representative hematoxylin and eosin (H&E)–stained colorectal histology patch with segmentation annotations produced by the proposed weakly supervised teacher–student framework. The color-coded contours indicate different glandular categories predicted by the model: red denotes malignant glands, green denotes benign glands, and blue indicates poorly differentiated clusters/glands (PDC/G). Regions without colored annotations correspond to background stroma. This example illustrates the framework's ability to identify and delineate multiple glandular structures across the tissue section under weak supervision. Color coding: red = malignant glands, green = benign glands, black = background stroma (best viewed in color).

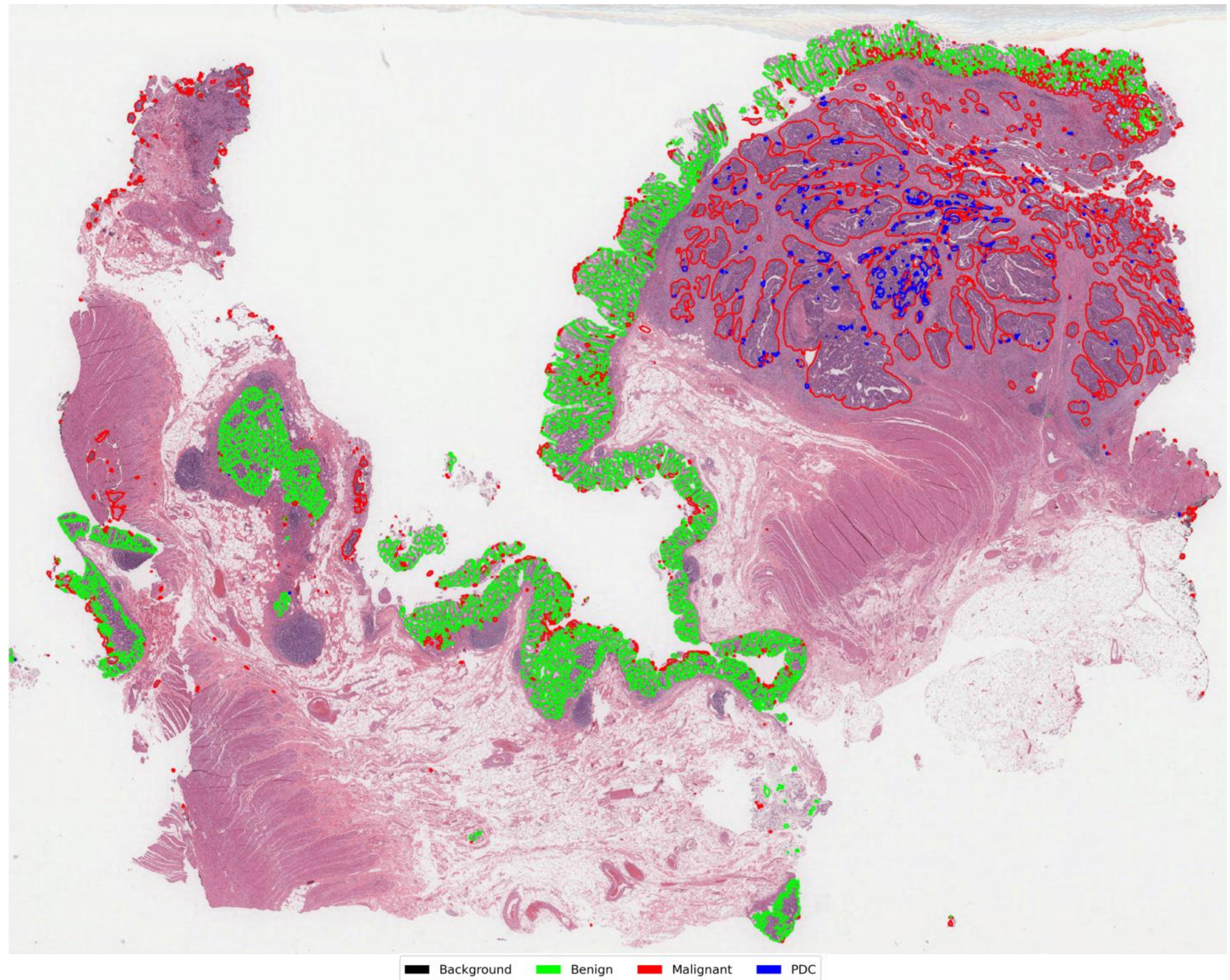

**Fig. 9.** Visualization of gland segmentation annotations generated by the proposed framework. A representative hematoxylin and eosin (H&E)–stained colorectal histology patch with segmentation annotations produced by the proposed weakly supervised teacher–student framework. The color-coded contours indicate different glandular categories predicted by the model: red denotes malignant glands, green denotes benign glands, and blue indicates poorly differentiated clusters/glands (PDC/G). Regions without colored annotations correspond to background stroma. This example illustrates the framework's ability to identify and delineate multiple glandular structures across the tissue section under weak supervision. Color coding: red = malignant glands, green = benign glands, black = background stroma (best viewed in color).

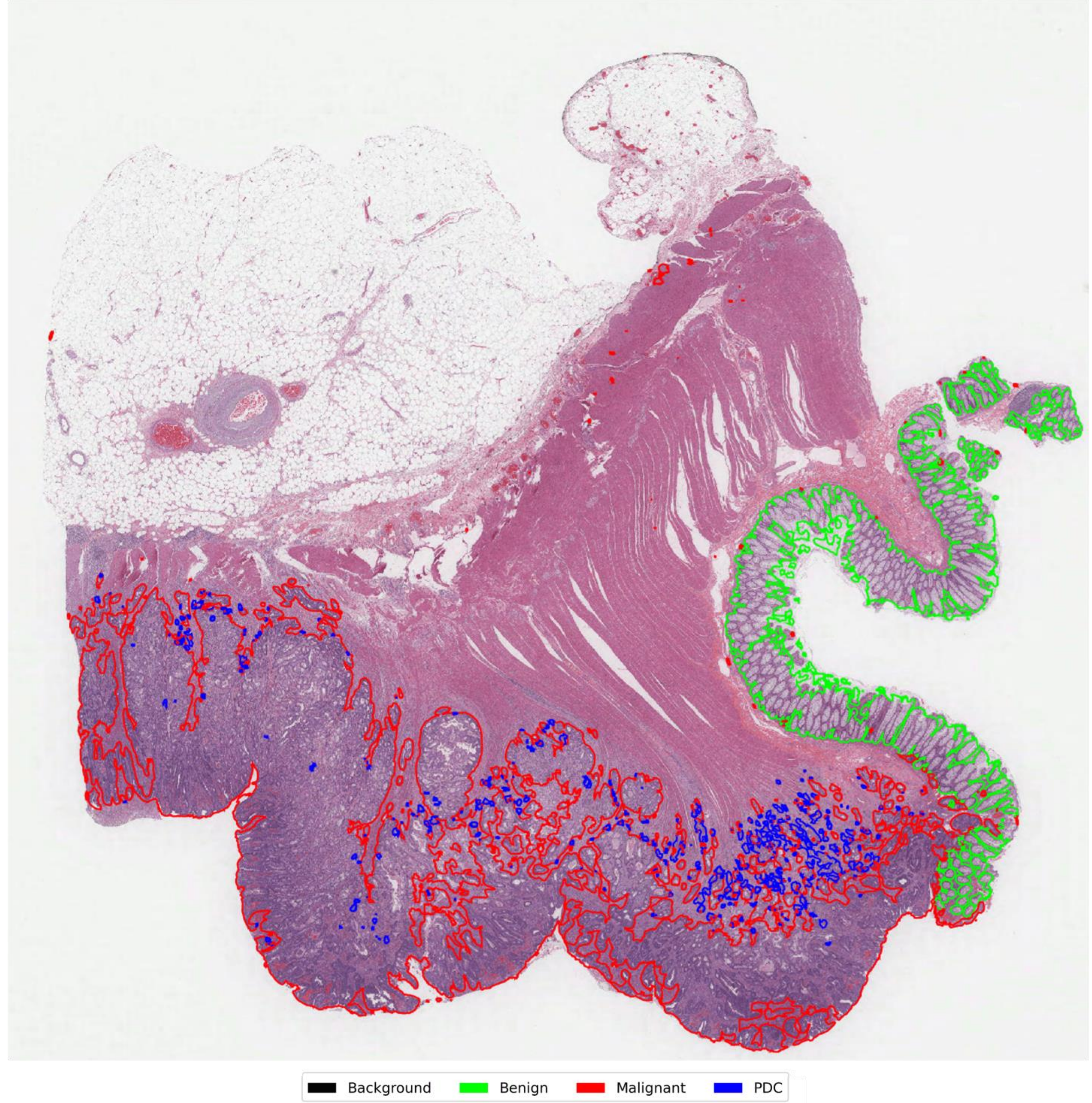

**Fig. 10.** Visualization of gland segmentation annotations generated by the proposed framework. A representative hematoxylin and eosin (H&E)–stained colorectal histology patch with segmentation annotations produced by the proposed weakly supervised teacher–student framework. The color-coded contours indicate different glandular categories predicted by the model: red denotes malignant glands, green denotes benign glands, and blue indicates poorly differentiated clusters/glands (PDC/G). Regions without colored annotations correspond to background stroma. This example illustrates the framework's ability to identify and delineate multiple glandular structures across the tissue section under weak supervision. Color coding: red = malignant glands, green = benign glands, black = background stroma (best viewed in color).

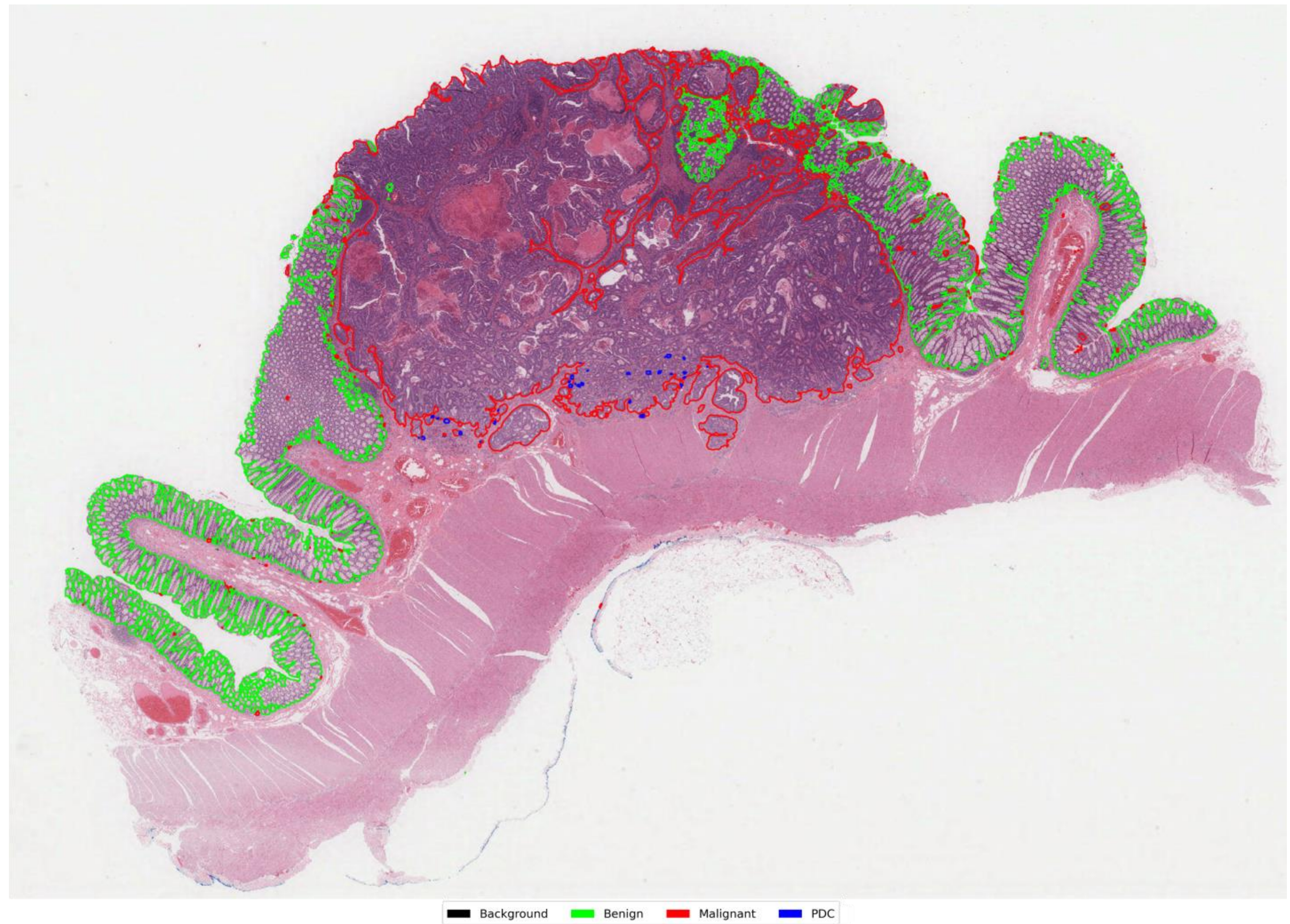

**Fig. 11.** Visualization of gland segmentation annotations generated by the proposed framework. A representative hematoxylin and eosin (H&E)–stained colorectal histology patch with segmentation annotations produced by the proposed weakly supervised teacher–student framework. The color-coded contours indicate different glandular categories predicted by the model: red denotes malignant glands, green denotes benign glands, and blue indicates poorly differentiated clusters/glands (PDC/G). Regions without colored annotations correspond to background stroma. This example illustrates the framework's ability to identify and delineate multiple glandular structures across the tissue section under weak supervision. Color coding: red = malignant glands, green = benign glands, black = background stroma (best viewed in color).

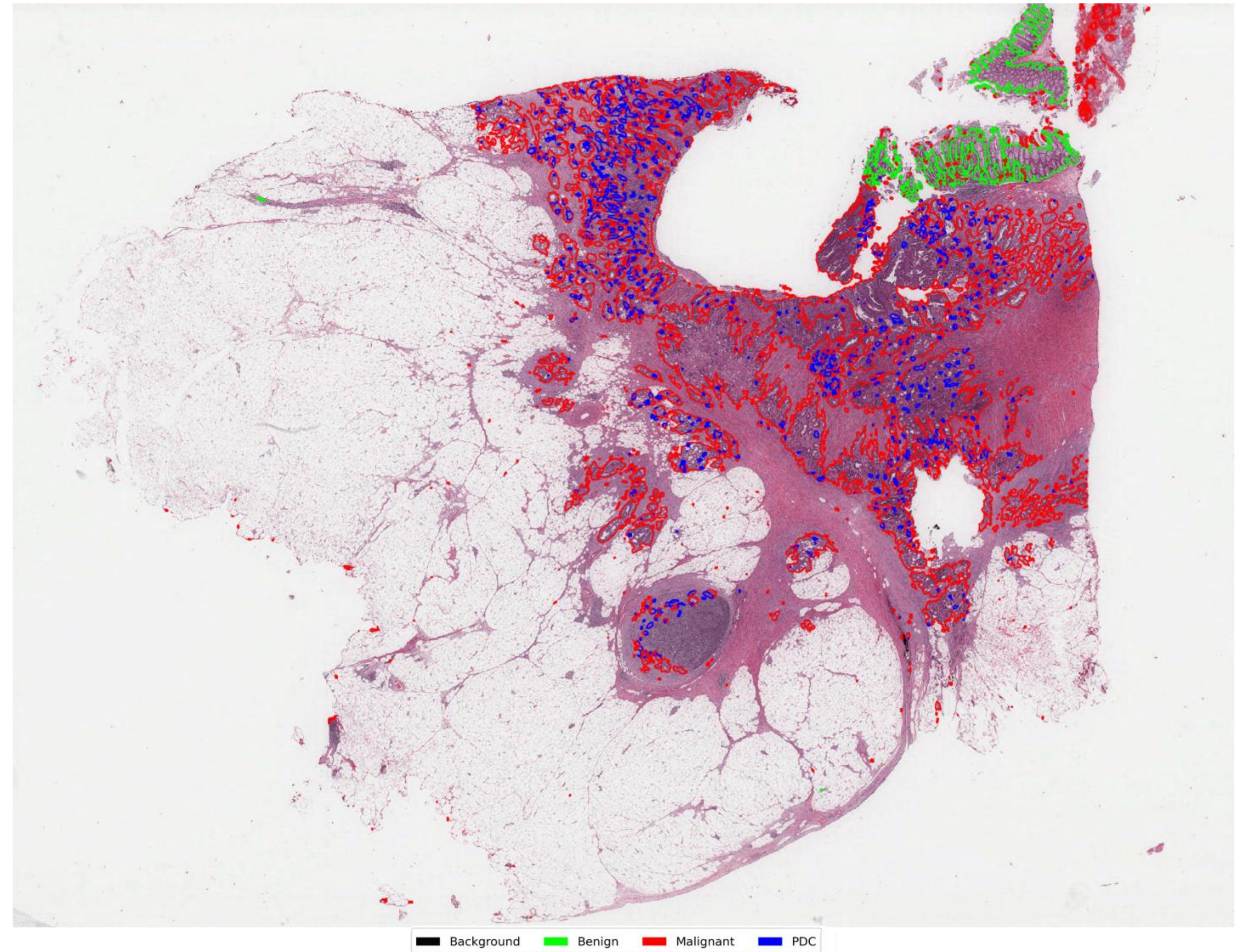

**Fig. 12.** Visualization of gland segmentation annotations generated by the proposed framework. A representative hematoxylin and eosin (H&E)–stained colorectal histology patch with segmentation annotations produced by the proposed weakly supervised teacher–student framework. The color-coded contours indicate different glandular categories predicted by the model: red denotes malignant glands, green denotes benign glands, and blue indicates poorly differentiated clusters/glands (PDC/G). Regions without colored annotations correspond to background stroma. This example illustrates the framework's ability to identify and delineate multiple glandular structures across the tissue section under weak supervision. Color coding: red = malignant glands, green = benign glands, black = background stroma (best viewed in color).

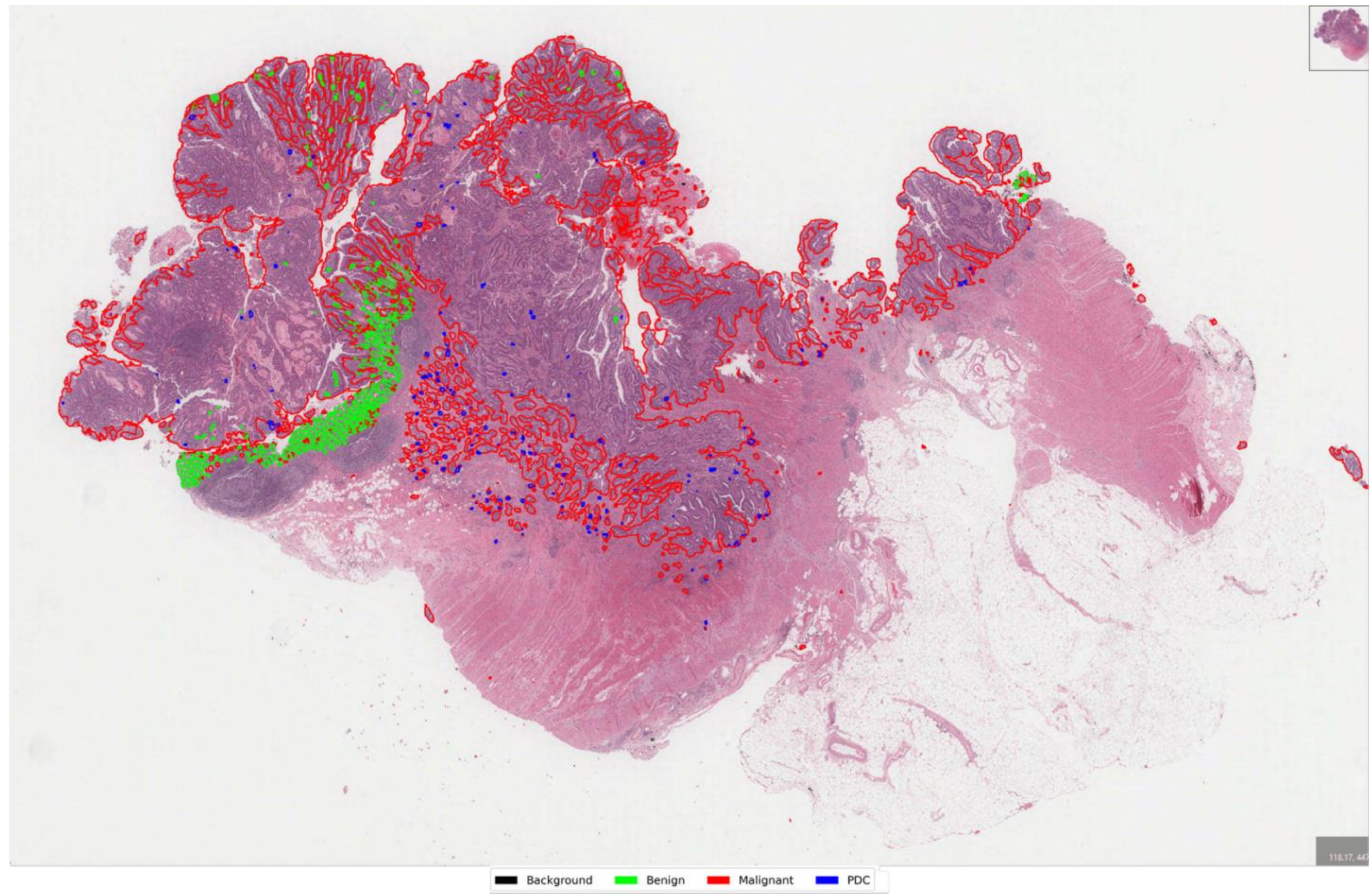

**Fig. 13.** Visualization of gland segmentation annotations generated by the proposed framework. A representative hematoxylin and eosin (H&E)–stained colorectal histology patch with segmentation annotations produced by the proposed weakly supervised teacher–student framework. The color-coded contours indicate different glandular categories predicted by the model: red denotes malignant glands, green denotes benign glands, and blue indicates poorly differentiated clusters/glands (PDC/G). Regions without colored annotations correspond to background stroma. This example illustrates the framework's ability to identify and delineate multiple glandular structures across the tissue section under weak supervision. Color coding: red = malignant glands, green = benign glands, black = background stroma (best viewed in color).

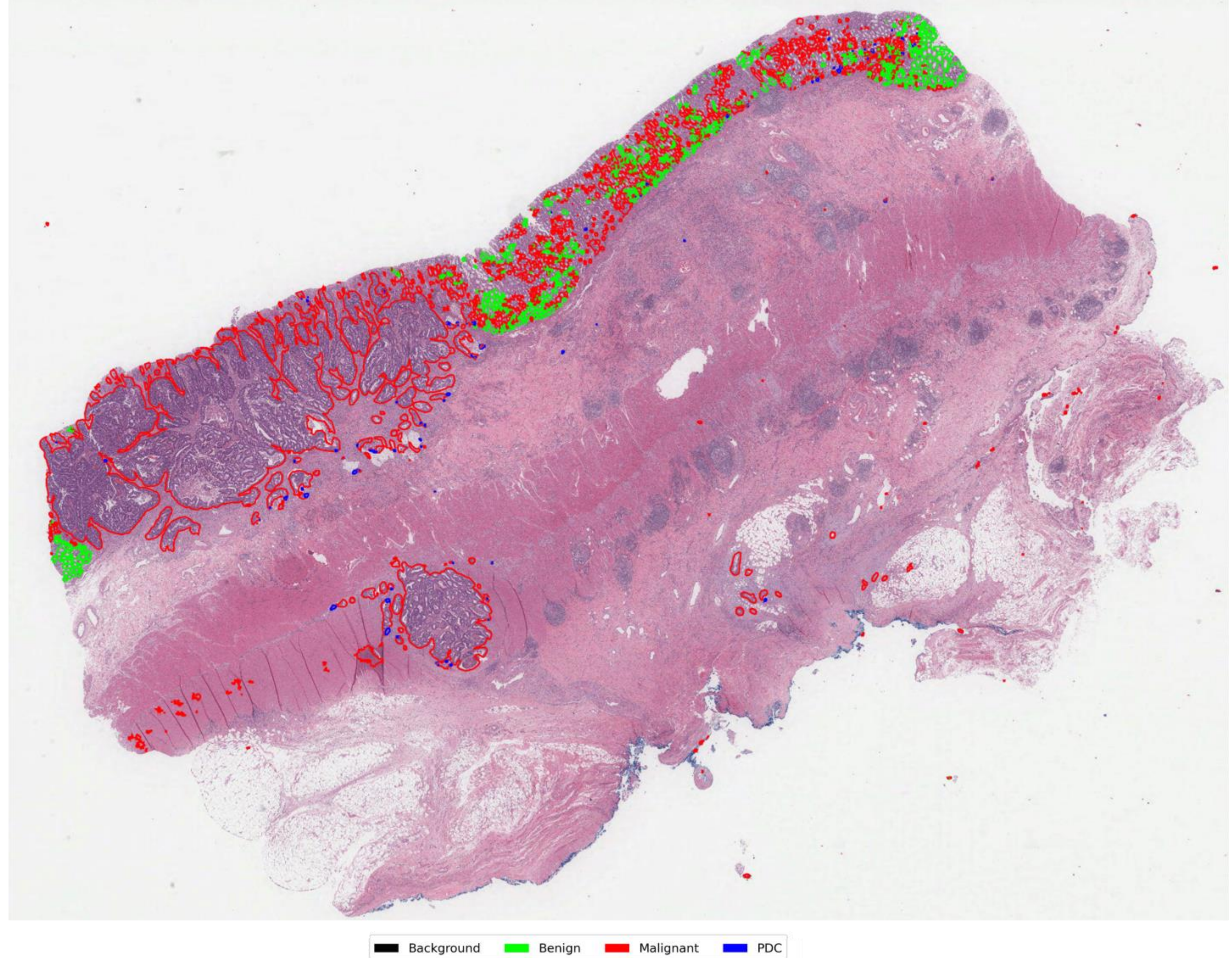

**Fig. 14.** Visualization of gland segmentation annotations generated by the proposed framework. A representative hematoxylin and eosin (H&E)–stained colorectal histology patch with segmentation annotations produced by the proposed weakly supervised teacher–student framework. The color-coded contours indicate different glandular categories predicted by the model: red denotes malignant glands, green denotes benign glands, and blue indicates poorly differentiated clusters/glands (PDC/G). Regions without colored annotations correspond to background stroma. This example illustrates the framework's ability to identify and delineate multiple glandular structures across the tissue section under weak supervision. Color coding: red = malignant glands, green = benign glands, black = background stroma (best viewed in color).